\documentclass[conference]{IEEEtran}
\IEEEoverridecommandlockouts

\usepackage[english]{babel}

\usepackage{cite}
\usepackage{graphicx}
\usepackage[cmex10]{amsmath}
\usepackage{multirow}
\usepackage{array}
\usepackage{tikz}
\usepackage[font=small]{caption}
\usepackage[lofdepth,lotdepth]{subfig}
\usepackage{marvosym}

\def\checkmark{\tikz\fill[scale=0.4](0,.35) -- (.25,0) -- (1,.7) -- (.25,.15) -- cycle;}

\newcommand{\shvspace}{\vspace{-2ex}}
\newcommand{\sshvspace}{\vspace{-2ex}}
\newcommand{\rhvspace}{\vspace{-1ex}}

\begin{document}

\small

\vspace{-2ex}

%
\title{Fusing Deep Convolutional Networks for Large Scale Visual Concept Classification}

\author{
\IEEEauthorblockN{Hilal Ergun and Mustafa Sert\textsuperscript{\Letter}}
Department of Computer Engineering\\
Ba\c{s}kent University\\
06810 Ankara, TURKEY\\
21020005@mail.baskent.edu.tr, \textsuperscript{\Letter}msert@baskent.edu.tr
}

\maketitle

\begin{abstract}
Deep learning architectures are showing great promise in various computer vision domains including image classification, object detection, event detection and action recognition. In this study, we investigate various aspects of convolutional neural networks (CNNs) from the big data perspective. We analyze recent studies and different network architectures both in terms of running time and accuracy. We present extensive empirical information along with best practices for big data practitioners. Using these best practices we propose efficient fusion mechanisms both for single and multiple network models. We present state-of-the art results on benchmark datasets while keeping computational costs at a lower level. Another contribution of our paper is that these state-of-the-art results can be reached without using extensive data augmentation techniques.
\end{abstract}

\IEEEpeerreviewmaketitle

\IEEEpubidadjcol

\shvspace
\section{Introduction}

Over the last few years, convolutional neural networks (CNNs) have demonstrated excellent performance in plethora of computer vision applications. Most of this success is mainly attributed to two factors; recent availability of parallel processing architectures and larger image datasets \cite{DBLP:journals/corr/ZeilerF13} \cite{simonyan2014very}. Recent arrival of annotated and bigger sized datasets like ILSVRSC \cite{deng2009imagenet}, LabelMe \cite{russell2008labelme} and MIT Places \cite{zhou2014learning}, made it possible to train deep convolutional networks for the task in hand without hitting the barrier of over-fitting. Furthermore, ability to access to higher computational resources allowed researchers to design deeper networks with more parameters.

One very important property of CNNs is their generalization ability. It is possible to apply a pre-trained network to a totally different dataset or application domain and achieve near state-of-the-art results. Moreover, with very little training effort they can beat most of other state-of-the-art approaches. This raises the question whether they can be applied to application domains where diversity of data is a challenge. Recent studies indeed show that this is perfectly possible \cite{razavian2014cnn}.

In addition to their transferability, CNNs are suitable for aggressive parallelization both in distributed environments and in high-performance GPU computing environments. Once trained, they can achieve running time speeds higher than real-time during inference. These attractive properties of CNNs make them a strong candidate for big multimedia applications. Whether they can be more than a candidate and become part of demanding applications depends on how they perform with next frontier of big data, i.e. video analysis applications. We are making this study to further investigate CNNs performances for big data applications. Our attempt is to reach out an implementation which is efficient in computing costs while maintaining best possible accuracy for the task in hand.

CNNs adapted to various image understanding tasks including but not limited to image classification, object detection, localization, human pose estimation, event detection, action recognition and so forth. They have gained much attention in recent years, although they were introduced in the late 80s \cite{6795724}. Krizhevsky et. al. applied CNNs to the task of image classification and demonstrated excellent results\cite{NIPS2012_4824}. Zeiler showed how CNN models can be further developed using visualization techniques \cite{DBLP:journals/corr/ZeilerF13}. In the following years deeper architectures surfaced further pushing classification accuracies. Winners of classification and localization tasks of ILSVRSC 2014 challenge \cite{deng2009imagenet} employed deeper architectures \cite{DBLP:journals/corr/SzegedyLJSRAEVR14} \cite{simonyan2014very}. Sermanet et. al. showed that CNNs can be trained to fulfill multiple image tasks like classification and localization together \cite{sermanet2013overfeat}. Latest studies reveal that much deeper architectures are expected to surface in the very near future\cite{he2015deep}.

Application of CNN architectures to video domain is also extensive. Karpathy et al. showed how CNN models can be enriched with temporal information present in videos using several fusion techniques \cite{karpathy2014large}. Simonyan et al. improved general approach to video action recognition using two stream convolutional networks \cite{simonyan2014two}. Ye and others further analyzed two stream architectures for video classification \cite{ye2015evaluating}. Wang et al. outlined good practices for action recognition in videos \cite{wang2015towards}. Zha et al. applied CNNs along with shallow architectures to video classification \cite{zha2015exploiting}.

As it has been shown in many previous work, implementation details play an important role in performance of vision algorithms. In this study, we compare recent state-of-the-art CNN implementations with similar processing pipelines with big data in mind. We aim to derive a set of best practices both concerning computational cost and accuracy constraints towards application of better big data analytic applications. Moreover, we analyze performance of 4 different popular CNN architectures introduced by previous studies. In addition to 4 CNN implementations, we further compare results of same CNN network trained on 2 different image datasets. We investigate different fusion strategies for incorporating multiple models and datasets. We also investigate effects of descriptor pooling strategies for efficient feature extraction. Last but not the least, we examine effects of applying CNNs to images at different scales. Our aim is to provide extensive empirical results for parameter selection strategies.

Rest of this paper is laid out as follows. In section 2 we provide detailed information about our visual object extraction methodologies. In section 3 we present our results and provide related discussion. We conclude in the last section with possible future research directions.

\shvspace
\section{Visual Concept Extraction Method}

Our proposed visual concept extraction scheme is depicted in Figure 1. Given an image or a video frame, proposed scheme passes given input through different network architectures in parallel. Depending on the concept extraction parameters  we extract several visual object features from several layers of different network architectures. We then encode extracted features using a feature fusion step. Feature fusion methodologies may differ, we mention those in section 2B. We describe details of each step in the following sections.

\begin{figure}[]
\centering
\includegraphics[scale=0.25]{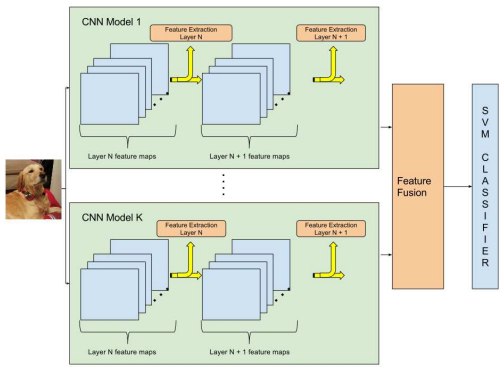}
\caption{Block diagram of visual concept extraction method}
\label{figure1}
\vspace{-1ex}
\end{figure}

\sshvspace
\subsection{CNN Architectures}
We investigate run-time performance of four frequently used CNN architectures. First one is the ground breaking work of Krizhevsky et.al introduced in 2012 for the image classification task of ILSVRSC challenge \cite{NIPS2012_4824}.

Second CNN model under investigation is the winner of 2014 ILSVRSC challenge by Szegedy et al. \cite{DBLP:journals/corr/SzegedyLJSRAEVR14}. Their network is a so called "Network-in-Network" architecture which was firstly introduced by Lin et al. \cite{DBLP:journals/corr/LinCY13}.

3rd and 4th models are the recently published work of Simonyan et al.\cite{simonyan2014very}, they are the runner-up of 2014 ILSVRSC classification challenge and the winner of same years localization task. These two convolution networks are very similar in design but one being 3 layers deeper than the other. When compared with other network architectures, they use smaller convolutional filters in all layers of the networks which permits them to increase their network depth.

In addition to aforementioned CNN architectures, we also evaluate another AlexNet model trained on MIT Places dataset\cite{zhou2014learning}. Use of this CNN model serves two purposes. One, we investigate fusion of two CNN models trained on different image datasets. Secondly, we further aim to show generalization capabilities of convolutional networks. Table ~\ref{table 2} shows our naming scheme of different networks through our evaluations. All the evaluations are conducted using Caffe framework \cite{jia2014caffe}.

\begin{table}[]
  \caption{CNN Architectures in Use}
  \centering
  \small
  \setlength\extrarowheight{3pt}
  \resizebox{0.6\columnwidth}{!}{%
  \begin{tabular}{|c|c|c|c|}
    \hline
    Network & Pre-Training & Scale & Code Name \\
    \hline
    AlexNet & ImageNet & 227x227 & M1 \\
    \hline
    AlexNet & ImageNet & 451x451 & M2 \\
    \hline
    VGG16 & ImageNet & 224x224 & M3 \\
    \hline
    VGG19 & ImageNet & 224x224 & M4 \\
    \hline
    GoogleNet & ImageNet & 224x224 & M5 \\
    \hline
    AlexNet & MIT Places & 227x227 & M6 \\
    \hline
    VGG16 & ImageNet & 448x448 & M7 \\
    \hline
  \end{tabular}%
  }
  \label{table 2}
  \vspace{-5ex}
\end{table}

\sshvspace
\subsection{Fusion Methodologies}

We investigate different fusion strategies on different aspects of our classification pipeline. One of the latest trends in architecture design is to merge different models into one modality. Many of the top performers in ILSVRSC 2016 challenge integrate this method into their classification pipeline. One can also merge information from different training datasets and we call this fusion scheme as dataset fusion. First we start with two identical CNN models which are trained on two distinct image datasets. One network is pre-trained on ImageNet and another is pre-trained on MIT Places dataset. Next we extract two different image descriptors using these models and apply SVM classification after feature fusion.

Another information fusion strategy can be such that different layers of a given CNN model can be used to extract different image features. Depending on the convolution and input size of a given network layer, one can extract total of \emph{M} spatially distinct feature maps with size $CxC$ from a given CNN layer. Later, feature vectors extracted from different layers can be fused using a simple concatenation operation. One drawback of this method is that it rapidly exploits feature size, especially  with networks which have low convolutional pitch. In order to overcome this cardinality problem we propose to use sum or max pooling for feature vectors which achieve state-of-the-art accuracy in our experiments with relatively low computational complexity. Figure 2 visualizes this fusion strategy. Mentioned fusion strategies can be attributed of being early fusion strategies since features are merged before they are introduced to SVM classifier. We also evaluate the use of late fusion strategy where classification decision outputs of multiple SVMs are merged to decide final classification decision.

\begin{figure}[]
\centering
\includegraphics[scale=0.37]{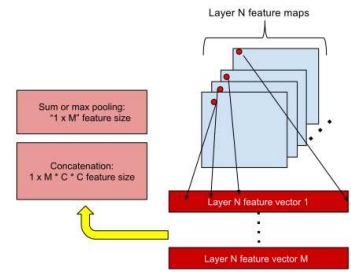}
\caption{Layer fusion methodologies}
\label{figure3}
\vspace{-2ex}
\end{figure}

\begin{figure}[]
\centering
\includegraphics[scale=0.3]{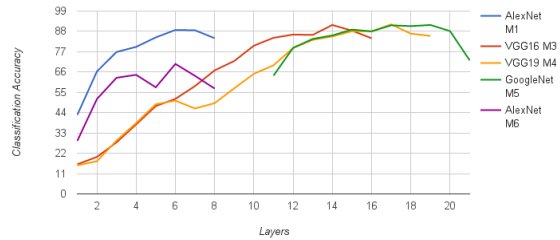}
\caption{Performance comparison of different layers on Caltech-101 Dataset for 5 CNN models}
\label{figure4}
\vspace{-4ex}
\end{figure}

\sshvspace
\subsection{Data Augmentation}

Data augmentation is a technique developed to prevent over-fitting problem during CNN training. Moreover, it is extensively used in testing phase as well. There are various augmentation techniques; including but not limited to RGB jittering \cite{NIPS2012_4824}, use of multiple crops, dense evaluation, image rotation and flipping. Their effect on classification accuracy is controversial. Almost all authors mention heavy usage of augmentation techniques in their papers while some reveal strong empiric information about marginality of their performance effects \cite{DBLP:journals/corr/SzegedyLJSRAEVR14}. Leaving this contradictory situation alone, we prefer to avoid use of data augmentation in this work for two different reasons. First of all, data augmentation, most of the time, requires passing a given image instance through CNNs multiple times which is not very efficient especially when we are dealing with big data. We believe that computational budget can be better spent running multiple CNNs in parallel while achieving higher accuracies and increased robustness. Secondly, data augmenting prevents us from comparing CNN results to shallow architectures like bag-of-words implementations utilizing vector quantization or sparse coding, Fisher vector or VLAD encoding. In previous attempts of the literature of image classification and object detection, data augmentation is rarely used \cite{chatfield2011devil} \cite{ergun2016efficient}.

One exception is multi-scale evaluation or dense evaluation. It is sometimes called scale jittering, and we use this technique in our work for various comparisons. Although it may be mainly classified as a data augmentation technique, it posses some unique features in the context of convolutions. First of all, applying convolutions at higher scales than the networks native working scale offers some attracting speed gains. For instance, in an architecture like AlexNet, doubling input resolution for each dimensions produces 64 output images at only 3x time. This permits us applying classification to one image in a densely fashion offering better classification accuracy compared to equivalent multiple crops application\cite{simonyan2014very}, with much shorter inference time. Secondly, it is the only way of applying CNNs to different input resolutions than the one trained for without utilizing spatial pyramid pooling layers, at least to our knowledge.

\sshvspace
\subsection{Datasets}

We report our results on 2 different image benchmarking datasets and 1 video action recognition dataset. We use Caltech-101\cite{fei2007learning} and Caltech-256\cite{griffin2007caltech} image benchmark datasets. Caltech-101 consists of 102 object categories spanning 9144 images. Caltech-256 contains 30607 images in 257 object categories. We also use UCF-101\cite{soomro2012ucf101} action recognition dataset for videos as it is one of the mostly used benchmark datasets in the literature and to better address the big data requirements. It contains clips of 101 action categories totaling 13320 videos of variable duration.

\shvspace
\section{Results}

We first investigate performances of single network layers on Caltech-101 dataset. Figure 3 summarizes results of this comparison. Scale jittered models are not shown in the graphic for simplicity, given that they are performing worse compared to their un-jittered counterparts. By looking at the results, we can clearly state that all models perform better on this dataset as we go deeper in network layers expect the final soft-max layers since they are tightly coupled with the pre-training dataset. These findings are consistent with the results presented in \cite{razavian2014cnn}.

We should note that in case of multi-dimensional convolution layers, we simply concatenate all feature maps of a given network layer into a single feature vector before passing layer data to the SVM classifier. Outputs of convolutional layers can be treated as feature maps generated by underlying receptive fields and they have higher dimensionality compared to final fully-connected layers. 4th column of Table ~\ref{table 3} shows dimensionality of some layers selected from different architectures. Use of any given layer alone is already sufficient to exploit feature cardinality for SVM classification, no need to mention features resulting from further multiple fusions. One solution to high cardinality problem is to use a suitable feature encoder like bag-of-words encoding or Fisher vector encoding or VLAD encoding. However, for computational purposes we choose a much simpler mechanism, pooling. We spatially pool feature maps either using maximum pooling or summation pooling. Choice of pooling is performed on a level dependent manner. As we go deeper in layers we observe that max pooling should be selected over sum pooling. 5th column of Table ~\ref{table 3} shows pooled feature sizes and it is clear that pooled layers has much lower feature cardinality than concatenated layers. Performance of different layers with different pooling types is visualized in Figure 5. Although pooling operations perform worse compared to concatenation of feature maps, their ability to keep feature dimensions at minimum is an indication of their suitability for feature fusion from multiple layers. We use max pooling in later sections for layer fusion operations.

\sshvspace
\subsection{Layer Fusion}

As a next step, we investigate fusion of different layers of a given CNN architecture. We choose 4 models for this setup; M1 (AlexNet), M2 (AlexNet scale jittered), M4 (VGG-19) and M5 (GoogleNet). To be fair with shallower architectures, we merge top 8 layers of each CNN architecture, however, layers are merged ground-up meaning when we merge 2 layers, we pick lowest convolutional layers from the top 8 layers of the corresponding architecture. This lets us to further comment on the performance of lower layers which are believed to capture lower level image semantics compared to deeper layers. We use max or sum pooling for convolutional layers to keep feature dimensions manageable as we inspected pooling performance on previous section. After we pool features of a given layer, we concatenate every layer into one feature vector and perform SVM classification. Table ~\ref{table 4} summarizes layer fusion results on Caltech-101 dataset. Last row of table belongs the best performance of single layer results of each CNN architecture. Merging multiple layers using simple pooling techniques clearly improves performance of any given network architecture. This is advantageous both from memory consumption and processing power perspectives with a reasonable gain in classification accuracy.

\begin{figure}[]
\centering
\includegraphics[scale=0.3]{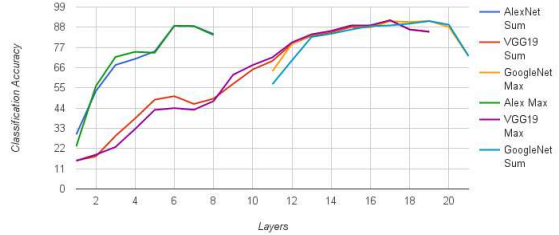}
\caption{Max pooling vs sum pooling for different CNN layers}
\label{figure5}
\end{figure}

\begin{table}[]
  \caption{Layer Dimensions of Various Architectures}
  \centering
  \small
  \setlength\extrarowheight{3pt}
  \resizebox{0.8\columnwidth}{!}{%
  \begin{tabular}{|c|c|c|c|c|}
    \hline
    Layer Name & Network & Dimensions & Linear Size & Max/Summed Size \\
    \hline
    Conv1 & AlexNet & 55x55x96 & 290400 & 96 \\
    \hline
    Conv2 & AlexNet & 27x27x256 & 186624 & 256 \\
    \hline
    conv4-1 & VGG19 & 28x28x512 & 401408 & 512 \\
    \hline
    conv5-1 & VGG19 & 14x14x512 & 100352 & 512 \\
    \hline
    inception-5a/output & GoogleNet & 7x7x832 & 40768 & 832 \\
    \hline
    inception-5b/output & GoogleNet & 7x7x1024 & 50176 & 1024 \\
    \hline
  \end{tabular}%
  }
  \label{table 3}
  \vspace{-3ex}
\end{table}

\begin{table}[t]
  \caption{Fusion Performance of Multiple Layers}
  \centering
  \small
  \setlength\extrarowheight{3pt}
  \resizebox{0.6\columnwidth}{!}{%
  \begin{tabular}{|c|c|c|c|c|}
    \hline
    \multirow{2}{*}{Fusion Count} & \multicolumn{4}{c|}{Classification Accuracy for Caltech-101} \\
    \cline{2-5}
    & M1 & M2 & M4 & M5 \\
    \hline
     8 Layers & \textbf{87.8326} & \textbf{84.8590} & \textbf{94.3333} &\textbf{ 94.1002} \\
    \hline
     7 Layers & 87.4024 & 85.5549 & 93.1826 & 92.9658 \\
    \hline
     6 Layers & 86.4024 & 83.6324 & 90.3846 & 92.7199 \\
    \hline
     5 Layers & 74.6696 & 73.9868 & 89.3402 & 91.0329 \\
    \hline
     4 Layers & 69.7840 & 69.8133 & 86.3190 & 89.8157 \\
    \hline
     3 Layers & 65.1558 & 62.6926 & 80.8952 & 89.3402 \\
    \hline
     2 Layers & 53.3187 & 50.5568 & 73.9342 & 87.8356 \\
    \hline
     1 Layer (Best) & 88.7699 & 83.5352 & 91.8062 & 91.4847 \\
    \hline
  \end{tabular}%
  }
  \label{table 4}
\end{table}

\sshvspace
\subsection{Model Fusion}

After layer fusion, we inspect effects of fusing different network architectures. This is one of the most frequent methods for improving classification accuracies, most of the participants of ILSVRC challenge submit fusions of different variants of their architectures. In this setup we choose six different network architectures, all pre-trained ImageNet dataset, and extract single layer features from all architectures. Here we choose best performing single layer of each network architecture. After feature extraction we apply either early or late fusion and finally classify resulting features using a linear SVM. Table ~\ref{table 5} summarizes our fusion results on benchmark datasets. We can clearly see that fusing different network models improves classification accuracy. Especially fusion of six different architectures achieves state-of-the-art results in all benchmark datasets.

\begin{table}[]
  \caption{Fusion performance of different network architectures}
  \centering
  \small
  \setlength\extrarowheight{3pt}
  \resizebox{0.8\columnwidth}{!}{%
  \begin{tabular}{|c|c|c|c|c|c|c|c|c|c|}
    \hline
    Dataset & M1 & M2 & M3 & M4 & M5 & M7 & Dim & Early & Late \\
    \hline
    Caltech-101 & \checkmark & \checkmark & \checkmark &  &  &  & 12288 & 94.2222 & - \\
    \hline
    Caltech-101 & \checkmark & \checkmark &  & \checkmark &  &  & 12288 & 94.2335 & 90.5156 \\
    \hline
    Caltech-101 & \checkmark & \checkmark & \checkmark & \checkmark &  &  & 16384 & 94.6862 & 91.9933 \\
    \hline
    Caltech-101 & \checkmark & \checkmark & \checkmark & \checkmark &  & \checkmark & 20480 & \textbf{95.0042} & 93.5082 \\
    \hline
    Caltech-101 & \checkmark & \checkmark & \checkmark & \checkmark & \checkmark & \checkmark & 21504 & \textbf{95.8024} & - \\
    \hline
    Caltech-256 & \checkmark & \checkmark &  & \checkmark &  &  & 12288 & 83.1259 & 78.354 \\
    \hline
    Caltech-256 & \checkmark & \checkmark &  & \checkmark &  & \checkmark & 16384 & 84.973 & 82.0237 \\
    \hline
    Caltech-256 & \checkmark & \checkmark &  & \checkmark & \checkmark & \checkmark & 17408 & 86.9478 & 84.1184 \\
    \hline
    UCF-101 & \checkmark & \checkmark & \checkmark &  &  &  & 12288 & 73.8832 & 66.2665 \\
    \hline
    UCF-101 & \checkmark & \checkmark & \checkmark & \checkmark &  &  & 16384 & 74.3061 & 69.0238 \\
    \hline
    UCF-101 & \checkmark & \checkmark & \checkmark & \checkmark &  & \checkmark & 20480 & 75.2313 & 71.8008 \\
    \hline
    UCF-101 & \checkmark & \checkmark & \checkmark & \checkmark & \checkmark & \checkmark & 21504 & \textbf{77.4782} & 73.2316 \\
    \hline
  \end{tabular}%
  }
  \label{table 5}
  \vspace{-3ex}
\end{table}

\sshvspace
\subsection{Dataset Fusion}

In this section, we investigate fusion of different CNN models trained on different datasets. We merge one CNN model trained on ImageNet with another CNN trained on MIT Places dataset. Although being marginal, these findings state that fusing multiple CNNs of different datasets helps to improve overall accuracy. We believe this further extends robustness of classifier against unseen data. Our dataset fusion results can be seen in Table ~\ref{table 6}. When compared to layer fusion and multiple model fusion, dataset fusion's effect on classification accuracy is rather marginal; this is especially the case for deeper architectures.

\begin{table}[t]
  \caption{Fusion of networks trained on different datasets}
  \centering
  \small
  \setlength\extrarowheight{3pt}
  \resizebox{0.8\columnwidth}{!}{%
  \begin{tabular}{|c|c|c|c|c|c|c|c|c|}
    \hline
    Dataset & M1 Only & M4 Only & M5 Only & M1+M6 & M4+M6 & M5+M6 \\
    \hline
    Caltech-101 & 88.7699 & 91.8062 & 91.4847 & 88.5199 & \textbf{92.2348} & \textbf{92.609}  \\
    \hline
    Caltech-256 & 70.6742 & 80.4877 & 79.6957 & \textbf{71.8542} & \textbf{80.8898} & \textbf{80.8031}  \\
    \hline
  \end{tabular}%
  }
  \label{table 6}
  \vspace{-2ex}
\end{table}

\sshvspace
\subsection{Action Recognition on Videos}

In order to better address the big data requirements, we evaluate our proposed scheme on real video data. We use action recognition dataset of UCF. We do not target temporal data in this study and present our results using only spatial information. To further challenge our proposed approach, we extract relatively low key-frames from a given video clip. We extract one frame at 2 seconds intervals from a given video frame which we believe is one of the most relaxed frame extraction intervals reported in the literature. Even though, we extract such low number of key frames from a given video our proposed scheme performs well under such low number of data and achieve near state-of-the-art results reported on UCF-101 dataset without using any temporal information. Last four rows of Table ~\ref{table 5} summarizes our findings.

\sshvspace
\subsection{Computational Complexity}

To shed some light on relatively dark side of CNN inference, we collected some run-time metrics during our tests. In Table ~\ref{table 7} we share absolute "frames per second" rating of each CNN architecture. Tests are conducted on a GTX980 GPU installed on relatively low-end i5-4460 CPU with 4 cores running on 3.20 GHZ clock speed. Rating are from end-to-end tests of each CNN model, i.e. from an image running through all layers of the corresponding architecture. Features are extracted from the last full-connected layers of all architectures. For the converted architectures, last convolutional layers sum pooled during feature extraction in order limit feature merging effects on performance.

\begin{table}[]
  \caption{Running Time Performance of CNN Architectures in Use}
  \centering
  \small
  \setlength\extrarowheight{3pt}
  \resizebox{0.6\columnwidth}{!}{%
  \begin{tabular}{|c|c|c|c|c|}
    \hline
    Network & Pre-Training & Scale & Code Name & FPS \\
    \hline
    AlexNet & ImageNet & 227x227 & M1 & \textbf{210} \\
    \hline
    AlexNet & ImageNet & 451x451 & M2 & 51 \\
    \hline
    VGG16 & ImageNet & 224x224 & M3 & 55 \\
    \hline
    VGG19 & ImageNet & 224x224 & M4 & 47 \\
    \hline
    GoogleNet & ImageNet & 224x224 & M5 & 58 \\
    \hline
    AlexNet & MIT Places & 227x227 & M6 & 210 \\
    \hline
    VGG16 & ImageNet & 448x448 & M7 & 15\\
    \hline
  \end{tabular}%
  }
  \label{table 7}
  \vspace{-4ex}
\end{table}

\sshvspace
\section{Conclusion}

We show that how CNN architectures can be used by preserving 2 Vs, namely the velocity and the value of big data in semantic information extraction from images and videos. We enhance single level CNN models by fusing information from different layers of a given CNN model. We reveal that simple averaging and maximizing pooling operations are able extract relevant information residing in different layers of network without hurting accuracy and can compete with other encodings which have relatively higher data cardinality. With the use of these best practices, it is possible to reach state-of-the-art results without using extensive data augmentation.

\rhvspace
\bibliographystyle{plain}
\bibliography{ieee_bigMM2016}

\end{document}